# An Adaptive Methodology for Ubiquitous ASR System

Urmila Shrawankar[1] & Vilas Thakare[2]

[1] G. H. Raisoni College of Engineering, Nagpur, India

[2] SGB Amravati University, Amravati, India

Correspondence: Urmila Shrawankar, G. H. Raisoni College of Engineering, Nagpur, India. E-mail: urmila@ieee.org

**Abstract**

Achieving and maintaining the performance of ubiquitous (Automatic Speech Recognition) ASR system is a real challenge.

The main objective of this work is to develop a method that will improve and show the consistency in performance of ubiquitous ASR system for real world noisy environment.

An adaptive methodology has been developed to achieve an objective with the help of implementing followings,

▪ Cleaning speech signal as much as possible while preserving originality / intangibility using various modified filters and enhancement techniques.

▪ Extracting features from speech signals using various sizes of parameter.

▪ Train the system for ubiquitous environment using multi-environmental adaptation training methods.

▪ Optimize the word recognition rate with appropriate variable size of parameters using fuzzy technique.

The consistency in performance is tested using standard noise databases as well as in real world environment. A good improvement is noticed.

This work will be helpful to give discriminative training of ubiquitous ASR system for better Human Computer Interaction (HCI) using Speech User Interface (SUI).

**Keywords:** ubiquitous ASR system performance, multi-environment adaptation training methods, real world environment, filters and enhancement techniques, ASR parameters optimization, fuzzy inference system for ASR, Human Computer Interaction (HCI), Speech User Interface (SUI)

## 1. Introduction

Speech User Interface (SUI) is a logical choice for man-machine communication, hence the growing interest in developing machines that accepts speech as input. Speech operated application in noisy environment is in demand, that is also very helpful to society for easy Human-Computer-Interaction.

However, a number of hurdles remain to make these technologies ubiquitous. In light of the increasingly mobile and socially connected population, core challenges include robustness to additive background noise, convolutional channel noise; room reverberation and microphone mismatch (IEEE Signal Processing Magazine, 2012). This so-called robustness problem not only leads to a significant degradation in performance but also hampers the fast commercialization of speech recognition applications.

Speech Recognition Systems give better results when the system is tested in conditions similar to the one used to train the acoustic models. It is very difficult to predict the noisy environment in advance in case of real world environmental noise and difficult to achieve environmental robustness.

Experimental results show that, a unique method is not available that will clean the noisy speech as well as preserve the quality which has been corrupted by real natural environmental (mixed) noise. It is also observed that, performance is depended on the parameters used while extracting speech signal features like, size of the window, frame, frame overlap etc. (Shrawankar & Thakare, 2012a)

The adaptation is a technique helps current recognition systems to solve this problem. An adaptive method presented in this paper uses variable size of parameters (window size, frame size and frame overlap percentage), various categories and levels of noise to train the system.

The method is developed (Shrawankar & Thakare, 2012b; Shrawankar & Thakare, 2010a) to clean noisy signals and enhanced them using two categories of techniques like traditional noise filters, and speech signal enhancement modified algorithms by considering all combination of enhancement techniques of three classes like Spectral Subtraction (Zhu, 2003), Subspace filtering (Ephraim & Trees, 1995), Statistical Filters (Lu & Loizou, 2010), independently as well as in combination.

Speech features are extracted form voiced signal (Shrawankar & Thakare, 2010b, Ramírez, Górriz, & Segura, 2007) using Mel-Frequency Cepstral Coefficients (MFCC) (Indrebo et al., 2008; Shrawankar & Thakare, 2010c) method.

A methodology is further developed for training all categories of noise that can adapt the acoustic models for a new environment that will help to improve as well as maintain the performance of the speech recognizer under real world environmental mismatched conditions. Training is done using Hidden Markov Models (HMM) (Sameti et al., 1998).

The analysis of performance is done using conventional as well as different objective (Ma et al., 2011; Ma et al., 2009; Hu et al., 2008) and subjective (Hu et al., 2006; Etame et al., 2011) measures that could be used to predict overall speech quality and speech/noise distortions introduced by representative speech enhancement algorithms from various classes (Shrawankar & Thakare, 2012b).

Performance of the system is tested for different categories of noise at various signal-to-noise ratio levels (Shrawankar & Thakare, 2013). Noise types include Airport, Car, Exhibition, Restaurant, Station, Street, Train, Factory, office, glass cabin etc.

Feasible and optimized sizes of parameters are optioned using fuzzy technique (Shrawankar & Thakare, 2012a).

The paper is organized as section 2 gives in detail of proposed Methodology followed by Empirical Process in section 3 and Results & Discussion in section 4 Concluding remarks are given in section 5.

## 2. Proposed Methodology

This complete experimental work focuses on following major issues for ubiquitous ASR performance improvement and maintaining consistency in performance.

▪ Cleaning speech signal: The first is speech signal filtering and enhancement for SNR improvement. The Indusial method and hybrid methods are implemented at back-end level and tested for the performance of the system with objective majors using SNR improvement test and subjective majors using listening test.

Following modified Filters and Enhancement techniques are used. Please refer cited papers for mathematical formulation.

- o Basic fundamental filters: Low-pass, High-pass, Band-pass and Band-shop.
- o Adaptive filters: Least Mean Squares (LMS) (Górriz & Ramírez, 2009), Adaptive Least Mean Squares (ALMS), Normalized Least Mean Squares filter (NLMS), Echo Return Loss Enhancement (ERLE).
- o Normalization techniques: RelAtive SpecTrAl (RASTA) (Hermansky & Morgan, 1994), CMN (Veth & Boves, 1996).
- o Enhancement methods: Spectral Subtraction – Boll (Boll, 1979), Spectral Subtraction – Berouti (Berouti, 1979), The Generalized Spectral Subtraction – Boh Lim Sim (Sim, 1998), Multi-band Spectral subtraction postriori – Kamath (Kamath et al., 2002), Wiener Filter a priori SNR – Wiener-Scalart (Breithaupt & Martin, 2010), MMSE-STSA – Ephraim (Ephraim & Malah, 1985), posteriori SNR – Ephraim and Malah (Ephraim & Malah, 1984), MMSE log-spectral Estimator – Cohen (Cohen, 2004), Kalman (Gannot et al., 1998) etc.

▪ Feature extraction: Extracting features from speech signals using various sizes of parameter (Zhu & Alwan, 2000). Five sets of features are extracted considering different size of frame, window and frame overlap.

▪ Training and testing using adaptive method: To train the system for all categories of environment, system uses ten different categories of environment speech samples, recorded at various locations. Five sets of parameters and ten categories of noise, total 50 training sets are used.

▪ Word recognition accuracy is calculated and checked the improvement at back-end as well as front-end level.

▪ Optimize the word recognition rate with appropriate variable size of parameters using fuzzy technique (Zadeh, 1965; Bezdek & Pal, 1992; Takagi & Sugeno, 1985): In this step a rule base Fuzzy Inference System (FIS) is used. SNR and World recognition accuracy are sent to the FIS as input parameters of all fifty set of features and best size of window, frame and frame overlap are computed for that category of noise as an output. Rules are framed to compute the output.

**3. Empirical Process**

A Software is prepared for the simulation. Experiment is performed with the help of following set of steps:

*Step 1: Samples Collection: Recording*

Recording is done outside at different locations mentioned above. Recording specifications are given in Table 1 and Table 2.

Table 1. Speech sample recording specifications

| | |
|---|---|
| Sampling Rate | 8 kHz |
| Time duration of each word sample | 3 Sec. |
| Recorded speech samples file format | WAV under Windows platform |
| Number of Speakers | 25 speakers (male and female) |
| Number of utterances | 5 utterances of each word from every speaker |
| Vocabulary type | Ten isolated words (number digits 0-9) |
| Recording Environment (10 different locations) | Airport, Car, Exhibition, Restaurant, Station, Street, Train, Factory, office, glass cabin (Clean) |

Table 2. Testing sample set specifications

| Noise Source | SNR Level | Type of Noise |
|---|---|---|
| NOIZEUS database (www.utdallas.edu/~loizou/speech/noizeus) | 20dB, 15dB, 10dB, 5dB, 0dB, -5dB. | Airport, Car, Exhibition, Restaurant, Station, Street, Train |
| NOISEX database www.speech.cs.cmu.edu/comp.speech/Section1/Data/noisex.html | unknown SNR level | Bubble, Buccaneer, Engine, Factory, Hfchannel, Machinegun, Pink, Volvo, White |
| Random Gaussian white noise | 20dB, 15dB, 10dB, 5dB, 0dB | Random |
| Real Environment (Natural) unknown noise | unknown SNR level | Locations: Airport, Car, Exhibition, Restaurant, Station, Street, Train, Factory, office, glass cabin |

*Step 2: Speech Signal Analysis*

Voice / Unvoiced/ Silent (VUS) Signal Identification:

The detection of the speech presence is calculated by detecting the beginning and end-point of an utterance using Voice Activity Detector (VAD) (Ramírez & Górriz, 2007). These two points detection algorithm is based on measures of the signal, zero crossing rate and short-time energy and checked whether the sample is voiced, unvoiced or silent. Only voiced samples are considered remaining samples are discarded.

*Step 3: Pre-Emphasis*

Under this Pre-emphasis step, Filters are implemented to estimate and reduce or filter the noise.

In order to illustrate the analysis of filtering and enhancement techniques fifty sets are considered.

The performance of the system is tested for all the considered combinations of the techniques. The noisy signals were filtered and enhanced using four categories of techniques like traditional noise filters for additive background noise; Adaptive filters for reducing reverberation effect, Normalization techniques for convolution noise and speech signal enhancement algorithms for clearing form distortion.

These filters and enhancement algorithms are implemented and tested for improving the intangibility of signal. The objective measures are checked by calculating the SNR and compared with SNR before implementing filter.

▪ Noise Filters

This category of filters is implemented for removing the noise from speech signals that are corrupted due to additive background noise.

In this experiment four fundamental traditional filters FIR like high-pass, low-pass, band-pass, and band-stop filters are implemented and tested. These filters are used for different frequency ranges, a high-pass filter for 20-22 Hz, a band-stop filter for 45-50 Hz and a low-pass filter for 3-4 KHz. Considering the energy of the signal, the speech is separated from noise.

▪ Adaptive Filtering

Room reverberation is also a one of the cause for speech signal distortion. Keeping this fact in mind, the system is tested using adaptive filters. These filters are implemented to improve the quality of speech signals those are distorted due to acoustic echo or reverberation. The quality improvement is tested with the help of adaptive filter algorithms like LMS, NLMS, ERLE, RLS etc.

▪ Normalization

The speech samples of words can be recorded using microphone, telephone device or any other recording instrument. There is a possibility that signals may get corrupted due to convolutional noise.

The normalization methods help to remove the convolutional noise originating from mismatches in microphone and/or channel characteristics, it is some form of speech enhancement.

This work uses RASTA (Hermansky & Morgan, 1994) and CMN (Veth & Boves, 1996) techniques to improve the performance.

*Step 4: Enhancement*

Speech enhancement algorithms attempt to recover a clean speech signal from a degraded signal containing additive noise. The evaluation of performance measures are performed using nine speech enhancement algorithms encompassing different classes such as spectral subtractive, signal subspace, statistical-model-based (MMSE, log-MMSE, and log-MMSE under signal presence uncertainty) and Wiener-filtering type algorithms (the a priori SNR estimation based method, the audible-noise suppression method are considered and tested for the performance.

*Step 5: Performance Evaluation*

Multiple methods are implemented independently as well as combinations of algorithms (Hybrid) to check the performance of a system. The performance evaluation is done on the basis of two performance measures, the first is objective evaluation using SNR improvement test and second is subjective quality evaluation is done using a informal listening test, spectrogram as well as waveform observation.

▪ Objective analysis (SNR improvement test)

The Signal-to-Noise Ratio (SNR) improvement test is considered as an objective measure (Ma et al., 2011; Ma et al., 2009; Hu et al., 2008) by calculating the SNR and is compared with SNR before implementing filter; spectrogram as well as waveform has been plotted and the clarity observed after implementing the filter or enhancement algorithm.

▪ Subjective analysis (Listening Test)

The subjective quality evaluation is done by using a listening test (Hu et al., 2006). The listening test is performed by normal hearing persons and the following parameters are observed.

- o   Overall quality (Intelligibility, Fidelity, Suppression etc)
- o   Musical noise salience , musical noise or other artifacts
- o   Preference

▪ Listening Test

Subsequent Informal listening tests are conducted for subjective evaluation. This test is a qualitative test. Ten volunteers were requested to evaluate the performance of the speech enhancement methods that were implemented in this project. The listeners gave their decisions on an individual basis. Ten speech samples were considered, each (digit) isolated word sample for every listener. First all the samples were numbered and played in the same order in which it was enhanced. The listeners ranked the methods based on the intelligibility and quality of the enhanced speech. On the basis of this test listener's observations are noted down.

*Step 6: Signal Enhancement using Hybrid Methods:*

As one of the aims of this work is to remove all categories of noise and distortion like additive noise, convolutional noise, reverberation etc. from the speech signal, the hybrid method is constructed. In the hybrid method, all enhancement methods are implemented with the combination of adaptive filters and normalization methods. Again the performance is observed using objective (SNR Improvement test) and subjective (informal listening test) parameters. The performance is tested for the proper combination of all categories of algorithms

*Step 7: Feature Extraction:*

Next important task is feature extraction. Signal is windowed with a specific window function (Hamming) using a window length, the word is partitioned into small units, called frames. The dimension of the frame is taken variable size from 10 ms to 50 ms, with 30-40% overlap. Feature extraction is executed for each frame independently. The spectrum is calculated for each window using the FFT. The spectrum is then filtered with a special Mel-scaled filter bank to get corresponding Mel-coefficients. The logarithm of Mel-coefficients is then computed. The discrete cosine transform is used to transform into the cepstrum-space. Non-necessary (high-frequency) MFCC-coefficients are discarded and finally 20 MFCC coefficients are considered. The extracted feature matrix (20 x 20) has been sent to train a model.

*Step 8: Training & Decoding:*

The feature vector obtained from MFCC is used to train the model. Training is giving using all types of samples, clean, artificial noise added, real world environmental noise and enhanced. For training models, the method applied is based on Hidden Markov Models (HMM). This system considers a Bakis model. Training procedure completed iteratively.

For the first iteration, random or equal (latter is default) numbers of frames are assigned to each state. The system uses the number of inputs equal to the number of coefficients extracted from a frame, and the number of outputs equal to the number of states of the model. The system is trained so that vectors of coefficients corresponding to each state activate the corresponding output. After training, the outputs can be interpreted as the values of emission PDF.

▪ Decoding

The next phase is decoding. Viterbi algorithm is applied and the best path is efficiently obtained, the path which has the highest probability. The probability is obtained from both emission and transition probabilities of the model. The value represents the probability of that model (with current parameters) corresponding to the observations. During this phase (training), the model is adjusted so that probability increases. Considering the best path, the correspondence between each frame and each state gets modified. First consequence is the modification of transition probabilities. Second consequence is the modification of the input vectors. The next iteration will begin with the new values for probabilities.

All categories of noisy samples are considered for the training.

*Step 9: Recognition*

After the system is trained, actual recognition begins. Given an unknown observation, determine which model generated it with more probability. Front-end analysis is applied and the coefficients are extracted. Then the probabilities of correspondence between each model and the observation are computed. This is done using Viterbi algorithm. The model with higher probability of compatibility is then recognized.

Word recognition accuracy is calculated using,

$$\text{Word Recognition Accuracy} = \frac{\text{Total Number of Words Recognised}}{\text{Total Number of Words tested}} \%$$

Word Recognition accuracy is tested for unknown as well as trained samples with 20% overlap.

*Step 10: Best Solution (Feasible and Optimised) finding using Fuzzy Inference System (FIS)*

While performing the experiment for evaluating the performance of speech processing methods, it is observed that every method behaves differently as parameter changes like hamming window size, frame size and overlapping size, filter used, enhancement algorithm implemented, category and type of noise etc. As it is very difficult to predict category of noise and implement proper variable size and algorithm for real world noisy environment.

Therefore it is desirable to obtain the best or optimized solution for these variabilities.

Finding the best variable size module uses a Rule-based Fuzzy Inference System. FIS is designed and computed best accuracy.

Fuzzy Approach is implemented with the help of five parts of the fuzzy inference process:

- Fuzzification of the input variables
- Application of the fuzzy operator in the antecedent
- Implication from the antecedent to the consequent
- Aggregation of the consequents across the rules
- Defuzzification

FIS is set using the following parameters:

Hamming Window, Frame Overlap percentage, Frame Size, SNR, Word Recognition Accuracy

```
[System]
Name='SpeechAccuracy'
Type='mamdani'
Version=2.0
NumInputs=3
NumOutputs=1
NumRules=5
AndMethod='min'
OrMethod='max'
ImpMethod='min'
AggMethod='max'
DefuzzMethod='centroid'
```

Three inputs are selected in the system, SNR value is passed for the Environment, Hamming windows size as WinSz and Frame overlap percentages as FrOver.

Input parameters, their membership function and ranges as follow.

```
[Input1]
```

Environment is defined as the value based on SNR, 10-20 dB is Very Noisy, 20-35 dB is Noisy and 35-50 dB is assumed for clean environment.

```
Name='Environment'
Range=[10 50]
NumMFs=3
MF1='VNoisy':'trimf',[-6 10 20]
MF2='Noisy':'trimf',[20 30 35]
MF3='Clean':'trimf',[35 50 66]
[Input2]
```

Window size is considered in three ranges Small, Medium and Large with ranges 240-250, 250-260 and 260-270 respectively.

Name='WinSz'

Range=[240 270]

NumMFs=3

MF1='Small':'trimf',[225 240 250]

MF2='Medium':'trimf',[250 255 260]

MF3='Large':'trimf',[260 270 282]

[Input3]

Frame overlap percentage is considered in three ranges Small, Medium and Large with ranges 20-40, 40-50 and 50-60 respectively.

Name='FrOver'

Range=[20 60]

NumMFs=3

MF1='Small':'trimf',[4 20 40]

MF2='Medium':'trimf',[40 45 50]

MF3='Large':'trimf',[50 60 76]

[Output1]

The Word recognition Accuracy is the final output. It is considered as Good, Better and Best in the expected range of 95 to 100 %,

Name='Accuracy'

Range=[95 100]

NumMFs=3

MF1='Good':'gaussmf',[0.8493 95]

MF2='Better':'gaussmf',[0.8493 97.5]

MF3='Best':'gaussmf',[0.8493 100]

[Rules]

After defining input, output and their membership functions, rules are framed and weights are assigned as given below

1. If (Environment is Clean) then (Accuracy is Better) (0.5)

2. If (Environment is Clean) and (FrOver is Medium) then (Accuracy is Best) (0.75)

3. If (Environment is Clean) and (WinSz is Medium) and (FrOver is Medium) then (Accuracy is Best) (1)

4. If (FrOver is Medium) then (Accuracy is Better) (0.5)

5. If (WinSz is Medium) then (Accuracy is Better) (0.5)

Final step is defuzzification, output accuracy is observed for different rules and crisp value is obtained using DefuzzMethod, centroid.

## 4. Results and Discussion

Very first, performance analysis of different filters and enhancement algorithms are done with the help of SNR improvement test and listening test, results are shown in Table 3 and comparative study is shown in Figure 1.

Table 3. SNR Improvement analysis using different enhancement algorithms

| Noise/Filter | SNR before Method | Boh | Boll | Berouti | Kamath | Wiener | Malah | Ephraim | Cohen | Kalman |
|---|---|---|---|---|---|---|---|---|---|---|
| Airport | 0.4676 | 2.1137 | 24.5838 | 25.0133 | 17.5450 | 29.7089 | 3.7800 | 1.4146 | 2.8227 | 13.6751 |
| Car | 0.4349 | 2.1033 | 24.9533 | 25.7274 | 17.5413 | 30.5197 | 4.0462 | 2.4220 | 2.3761 | 14.0251 |
| Exhibition | 0.4387 | 2.0981 | 24.1546 | 23.6696 | 17.9683 | 28.1462 | 3.7704 | 1.5830 | 2.5505 | 12.9532 |
| Restaurant | 0.3925 | 2.0974 | 25.1374 | 25.6120 | 17.5581 | 29.9062 | 3.9725 | 1.4003 | 1.5021 | 15.1954 |
| Station | 0.4473 | 2.0965 | 25.0341 | 25.6080 | 17.5900 | 29.7828 | 4.1638 | 1.3836 | 0.8602 | 13.3720 |
| Street | 0.4791 | 2.1037 | 25.0670 | 25.3666 | 17.6736 | 30.1571 | 4.0342 | 2.4239 | 2.4571 | 14.3421 |
| Train | 0.6334 | 2.1086 | 24.6206 | 25.6357 | 17.1553 | 29.5159 | 3.8546 | 1.3915 | 1.0772 | 13.2642 |
| Office | 2.5320 | 2.0871 | 23.3656 | 25.1451 | 16.7053 | 29.0016 | 4.0662 | 1.3362 | 4.3488 | 15.5965 |
| Cabin | 2.2710 | 2.0932 | 24.5531 | 25.7961 | 17.0794 | 29.4610 | 4.3504 | 1.6457 | 1.6216 | 14.6930 |
| Factory | 0.0299 | 2.0866 | 24.7391 | 25.7618 | 17.4127 | 28.7751 | 4.4487 | 1.5160 | 1.2652 | 13.9924 |

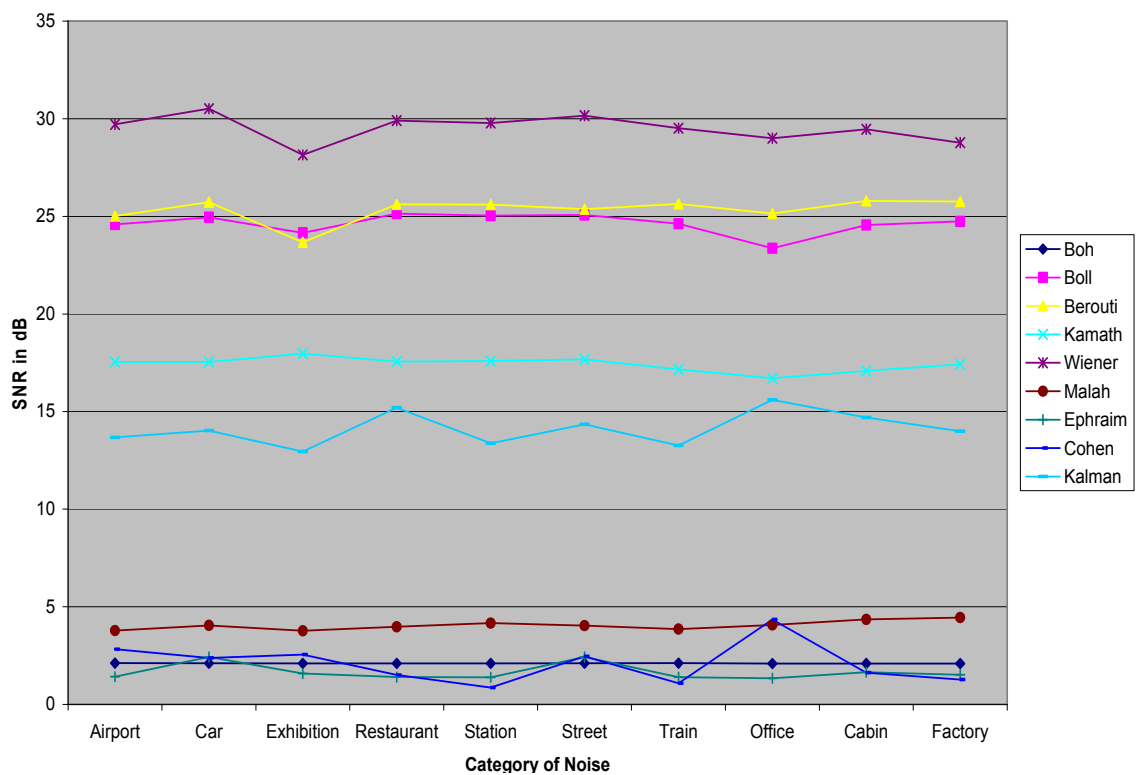

Figure 1. Analysis of enhancement methods

Result shows spectral subtraction and wiener filters are suitable techniques for removing mixed noise. Further it is observed that hybrid methods (combination of adaptive filters, reverberation filters and enhancement method) improves the SNR value (shown in Table 4 and Figure 2) and helps in achieving better accuracy.

Table 4. SNR Improvement analysis using different hybrid methods

| Noise/Filter | SNR Before Method | LMS | RASTA | Boh | Boll | Berouti | Kamath | Wiener | Malah | Ephraim | Cohen | Kalman |
|---|---|---|---|---|---|---|---|---|---|---|---|---|
| Airport | -0.5798 | 0.0329 | 24.0722 | 27.9301 | 27.7881 | 19.2034 | 21.5547 | 31.8449 | 21.0915 | 0.0465 | 6.5477 | 17.4962 |
| Car | -0.5894 | 0.0353 | 24.1715 | 27.9273 | 28.5729 | 19.7936 | 21.6816 | 32.1340 | 11.0392 | 0.0648 | 6.2240 | 18.8437 |
| Exhibition | -0.5319 | 0.0353 | 24.0090 | 27.8513 | 25.0844 | 17.6295 | 21.4561 | 31.6925 | 17.2397 | 0.0612 | 6.5664 | 20.3842 |
| Restaurant | -0.6365 | 0.0349 | 24.1076 | 27.8150 | 28.4147 | 20.5257 | 21.6613 | 32.1850 | 10.5838 | 0.0676 | 4.0126 | 21.3422 |
| Station | -0.5691 | 0.0324 | 24.1194 | 27.7380 | 28.7481 | 19.8335 | 21.6629 | 32.0832 | 14.1524 | 0.0634 | 6.6996 | 19.4269 |
| Street | -0.6146 | 0.0359 | 24.1943 | 27.9013 | 27.7711 | 19.9870 | 21.6739 | 32.1313 | 11.6933 | 0.0513 | 6.6325 | 20.2453 |
| Train | -0.4772 | 0.0412 | 24.0577 | 27.9368 | 29.8063 | 19.9039 | 21.7620 | 31.9279 | 15.1135 | 0.0664 | 5.7845 | 16.3425 |
| Office | 7.7565 | 0.0540 | 24.0705 | 28.0360 | 21.2363 | 14.7459 | 22.9408 | 31.7663 | 18.6507 | 0.0603 | 5.6517 | 21.8258 |
| Cabin | 0.7583 | 0.4922 | 24.4010 | 27.7709 | 28.2481 | 18.4967 | 21.4113 | 32.0866 | 21.8687 | 0.0676 | 5.7008 | 18.4237 |
| Factory | -4.0913 | 0.0458 | 24.1316 | 27.9218 | 27.0322 | 18.8081 | 21.6833 | 31.9832 | 17.9429 | 0.0686 | 5.6940 | 17.3862 |

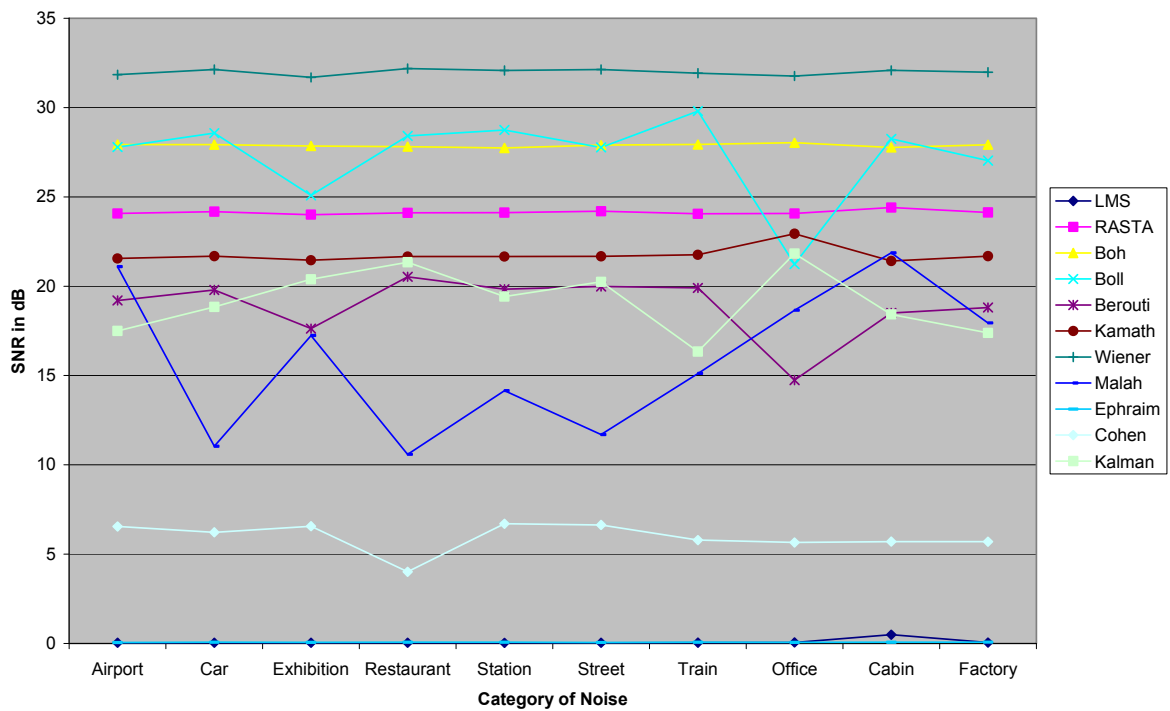

Figure 2. Analysis of hybrid methods

Accuracy is computed for different variable size of Hamming Window, Frame and % of Frame, results are shown in Table 5 for one sample and accuracy analysis is shown in Figure 3. Same experiment is performed for different samples collected from different locations.

Table 5. Accuracy analysis with variable size of hamming window, frame and % of frame overlap

| Hamm Win Size | Variables | Frame Overlap 20% | Frame Overlap 25% | Frame Overlap 30% | Frame Overlap 35% | Frame Overlap 40% | Frame Overlap 45% | Frame Overlap 50% | Frame Overlap 55% | Frame Overlap 60% |
|---|---|---|---|---|---|---|---|---|---|---|
| 240 | FrSz | 11.00 | 11.00 | 12.00 | 13.00 | 13.00 | 13.00 | 14.00 | 15.00 | 15.00 |
|  | SNR | 10.62 | 13.95 | 18.19 | 21.91 | 29.03 | 37.68 | 42.98 | 33.39 | 25.59 |
|  | Accuracy | 95.62 | 97.66 | 96.38 | 92.00 | 95.81 | 96.09 | 97.15 | 96.82 | 97.23 |
| 245 | FrSz | 11.00 | 11.00 | 12.00 | 12.00 | 13.00 | 13.00 | 14.00 | 14.00 | 15.00 |
|  | SNR | 9.64 | 13.11 | 17.75 | 22.22 | 29.58 | 39.16 | 43.83 | 33.07 | 23.87 |
|  | Accuracy | 86.78 | 91.80 | 94.09 | 93.32 | 97.61 | 99.87 | 99.06 | 95.91 | 90.69 |
| 250 | FrSz | 11.00 | 11.00 | 12.00 | 12.00 | 13.00 | 13.00 | 14.00 | 14.00 | 15.00 |
|  | SNR | 10.18 | 13.91 | 17.68 | 23.50 | 29.58 | 38.84 | 43.45 | 33.31 | 23.57 |
|  | Accuracy | 91.65 | 97.35 | 93.71 | 98.68 | 97.61 | 99.05 | 98.21 | 96.59 | 89.55 |
| 255 | FrSz | 10.00 | 11.00 | 11.00 | 12.00 | 12.00 | 13.00 | 13.00 | 14.00 | 14.00 |
|  | SNR | 10.09 | 13.97 | 16.94 | 22.21 | 29.00 | 36.11 | 39.36 | 30.90 | 23.69 |
|  | Accuracy | 90.78 | 97.79 | 89.77 | 93.29 | 95.71 | 92.08 | 98.95 | 98.60 | 90.01 |
| 260 | FrSz | 10.00 | 11.00 | 11.00 | 12.00 | 12.00 | 13.00 | 13.00 | 13.00 | 14.00 |
|  | SNR | 10.75 | 13.57 | 18.05 | 22.08 | 29.02 | 38.02 | 43.34 | 33.52 | 25.91 |
|  | Accuracy | 96.73 | 95.01 | 95.65 | 92.72 | 95.77 | 96.96 | 97.94 | 97.21 | 98.47 |
| 265 | FrSz | 10.00 | 11.00 | 11.00 | 11.00 | 12.00 | 12.00 | 13.00 | 13.00 | 14.00 |
|  | SNR | 10.34 | 13.78 | 16.88 | 21.99 | 28.97 | 37.21 | 41.90 | 33.21 | 25.57 |
|  | Accuracy | 93.04 | 96.46 | 89.48 | 92.36 | 95.61 | 97.88 | 98.70 | 96.30 | 97.18 |
| 270 | FrSz | 10.00 | 10.00 | 11.00 | 11.00 | 12.00 | 12.00 | 13.00 | 13.00 | 14.00 |
|  | SNR | 10.27 | 13.61 | 16.85 | 22.00 | 28.37 | 36.29 | 40.09 | 32.27 | 24.40 |
|  | Accuracy | 92.46 | 95.30 | 89.31 | 92.42 | 93.63 | 98.53 | 98.61 | 98.59 | 92.71 |

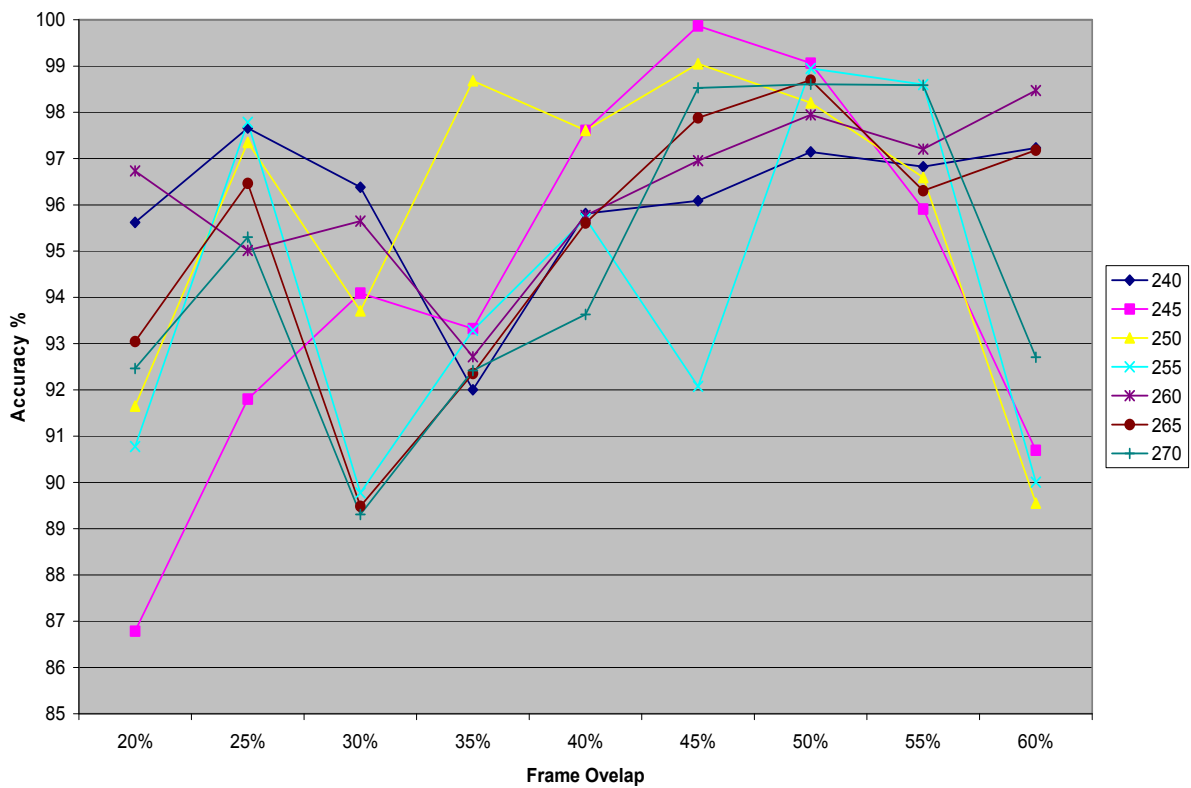

Figure 3. Word recognition accuracy for different size of window, frame and their overlap

Finally, these values are send to fuzzy rule based system and optimized size of variables are computed.

## 5. Conclusions

An Adaptive Methodology is very essential to improve the performance of ubiquitous ASR system as adverse environmental effects are not constant.

Adaptation is achieved by multi-environmental training with all probable combinations of variable sizes of window and frame etc while extracting features.

It is observed that hamming window size 245-250 ms and frame overlap 45% give best accuracy for ubiquitous ASR system.

As speech signal is cleaned with all possible hybrid methods, adverse environmental effects are normalized and hence environmental robustness is achieved.